\documentclass[runningheads]{llncs}
\usepackage{amssymb}
\usepackage{graphicx}
\graphicspath{{figures/}}
\usepackage{subfig}
\usepackage{amsmath}
\usepackage[misc]{ifsym} 
\usepackage{cite}
\usepackage{bm}
\begin{document}
\title{A Feedback Neural Network for Small Target Motion Detection in Cluttered Backgrounds}
\titlerunning{A Feedback Neural Network for Small Target Motion Detection}
\author{Hongxin Wang\inst{1} \and Jigen Peng\inst{2} \and Shigang Yue\inst{1}}
\authorrunning{Hongxin Wang \and Jigen Peng \and Shigang Yue}
\institute{The Computational Intelligence Lab (CIL), School of Computer Science, \\ University of Lincoln, Lincoln, LN6 7TS, UK \\
	\email{syue@lincoln.ac.uk} 
	\and School of Mathematics and Information Science, Guangzhou University, Guangzhou, 510006, China \\
	\email{jgpeng@gzhu.edu.cn} 
}

\maketitle              % typeset the header of the contribution
\begin{abstract}
Small target motion detection is critical for insects to search for and track mates or prey which always appear as small dim speckles in the visual field. A class of specific neurons, called small target motion detectors (STMDs), has been characterized by exquisite sensitivity for small target motion. Understanding and analyzing visual pathway of STMD neurons are beneficial to design artificial visual systems for small target motion detection. Feedback loops have been widely identified in visual neural circuits and play an important role in target detection. However, if there exists a feedback loop in the STMD visual pathway or if a feedback loop could significantly improve the detection performance of STMD neurons, is unclear. In this paper, we propose a feedback neural network for small target motion detection against naturally cluttered backgrounds. In order to form a feedback loop, model output is temporally delayed and relayed to previous neural layer as feedback signal. Extensive experiments showed that the significant improvement of the proposed feedback neural network over the existing STMD-based models for small target motion detection.

\end{abstract}
\section{Introduction}
In dynamic visual world, the observer (an animal) are more interested in moving objects, since they are more likely to be mates, predators or prey. Being able to detect moving objects in a distance and early could endow the observer with stronger competitiveness for survival. However, when an object is far away from the observer, it often appears as a small dim speckle whose size may vary from one pixel to a few pixels in the visual field. Detecting such small targets in visual cluttered backgrounds has been considered as a challenging problem for artificial visual systems. This is not only because shape, color and texture information of small targets cannot be used for motion detection, but also because the cluttered background, such as bushes, trees and/or rocks, always contains a great number of small-target-like features (called background noise). Small target motion detection means detecting small moving targets, meanwhile discriminating them from background noise.

Insects exhibit exquisite sensitivity for small target motion ~\cite{nordstrom2006insect} and can pursue small flying targets, such as mates or prey, with high capture rates \cite{olberg2000prey}. As revealed in biological research ~\cite{nordstrom2006insect,nordstrom2012neural}, the exquisite sensitivity is coming from a class of specific neurons in the insects' visual system, called small target motion detectors (STMDs). STMD neurons give peak responses to targets subtending $1-3^{\circ}$ of the visual field, with no response to larger bars (typically $> 10^{\circ}$) or to wide-field grating stimuli. The electrophysiological knowledge about STMD neurons and their afferent pathways is helpful for designing artificial visual systems for small target motion detection.

A few STMD-based models have been proposed for detecting small target motion in naturally cluttered backgrounds. Elementary small target motion detector (ESTMD) which was proposed by Wiederman \emph{et al.} \cite{wiederman2008model}, can detect the presence of small moving targets, but not the motion direction. To detect small moving targets and their motion directions, three directionally selective models have been proposed, including EMD-ESTMD \cite{wiederman2013biologically,bagheri2017performance}, ESTMD-EMD \cite{wiederman2013biologically,bagheri2017performance} and directionally selective small target motion detector (DSTMD) \cite{wang2018directionally}. Although these existing STMD-based models can detect small moving targets, their detection results often contain a great number of background noise. Further improvement is needed for filtering out background noise.

Feedback loops exist extensively in animals' visual systems and can optimize motion estimation \cite{kafaligonul2015feedforward,clarke2017feedback}. Biological research reveals that feedback loops are able to simultaneously mediate the synthesis of motion representations and cancellation of distracting signals \cite{clarke2017feedback}. However, it is still unclear if a feedback loop exist in the visual pathway of STMD neurons or if a feedback loop can significantly improve detection performance of STMD neurons. In this paper, we investigate that if a feedback loop exists, can it improve detection performance of STMD neurons. To answer this question, we propose a feedback neural network ({\bf feedback ESTMD}) based on the existing ESTMD model \cite{wiederman2008model} for small target motion detection. In order to form a feedback loop, model output is firstly temporally delayed and then relayed to previous neural layer (medulla layer) as feedback signal. The feedback signal is added on the output of medulla layer for weakening responses to background noise. Systematic experiments demonstrate that the feedback loop can significantly improve detection performance of the existing STMD-based models. 

The remainder of this paper is organized as follows. In Section \ref{Formulation-of-the-model}, the proposed feedback neural network is introduced in details. In Section \ref{Results-and-Discussions}, experiments are carried out to test the performance of the proposed feedback neural network. Discussion is also given in this section. In Section \ref{Conclusion}, we give conclusions and perspectives.

\section{Formulation of the Model}
\label{Formulation-of-the-model}

\begin{figure}
	\centering
	\includegraphics[width=0.95\textwidth]{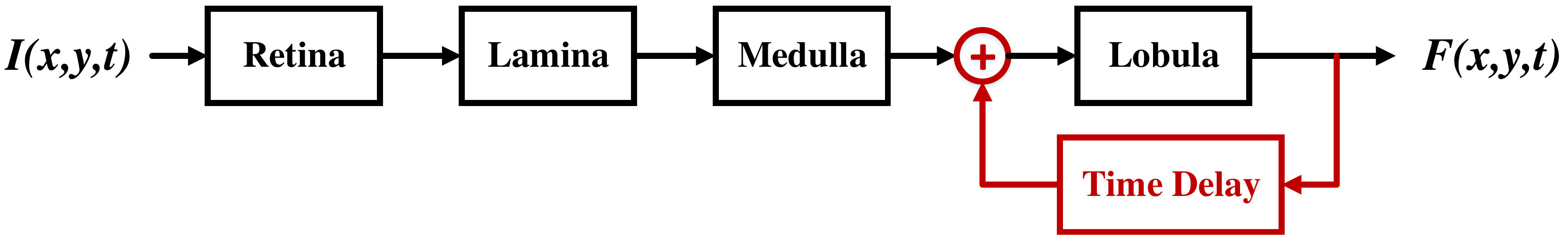}
	\caption{Schematic illustration of the proposed feedback model.}
	\label{Schematic-of-Feedback-Model}
\end{figure}

In this section, we elaborate on the proposed feedback model, called {\bf Feedback ESTMD}. Its schematic illustration is shown in Fig. \ref{Schematic-of-Feedback-Model}. As can be seen, $I(x,y,t)$ is the model input, denoting an image sequence where $x,y$ and $t$ are spatial and temporal field positions, respectively. Model input $I(x,y,t)$ is successively processed by four neural layers including retina, lamina, medulla and lobula. Through the process of four neural layers, we can obtain a model output $F(x,y,t)$. The output $F(x,y,t)$ is firstly temporally delayed and then relayed to medulla layer so as to form a feedback loop. The proposed feedback loop can weaken responses to background noise and significantly improve detection performance. In the following, functionalities of four neural layers and the feedback loop will be introduced in details.

\subsection{Retina Layer}

In the insect's visual system, retina layer contains a great number of ommatidia \cite{warrant2016matched}. These ommatidia are able to receive luminance signals from the natural world and relay signals to downstream neurons for further process. The received luminance signal are always highly blurred, due to the extremely low resolution of ommatidia.

In the proposed feedback neural network, each ommatidium is modeled as a spatial Gaussian filter for simulating ommatidium's blur effect. Let $I(x,y,t) \in \mathbf{R}$ denote the input image sequence where $x,y$ and $t$ are spatial and temporal field positions. Then, the output of ommatidium with visual field centered at $(x,y)$ denoted by $P(x,y,t)$ is defined as,
\begin{equation}
P(x,y,t) =  \iint I(u,v,t)G_{\sigma_1}(x-u,y-v)dudv
\label{Photoreceptors-Gaussian-Blur}
\end{equation}
where $G_{\sigma_1}(x,y)$ is a Gaussian function, given by
\begin{equation}
G_{\sigma_1}(x,y)= \frac{1}{2\pi\sigma_1^2}\exp(-\frac{x^2+y^2}{2\sigma_1^2}).
\label{Photoreceptors-Gauss-blur-Kernel}
\end{equation}

\subsection{Lamina Layer}
In the insect's visual system, lamina layer contains a great number of large monopolar cells (LMCs)  \cite{behnia2014processing}. LMCs receive signals from ommatidia and are able to extract motion information from ommatidium output. To be more precise, LMCs show strong responses to brightness increments and decrements, i.e., luminance changes. 

In the proposed feedback neural network, each LMC is modeled as a temporal high-pass filter extracting luminance changes, i.e., motion information,  from ommatidium output $P(x,y,t)$. Let $L(x,y,t)$ denote the output of LMC located at $(x,y)$. Then, $L(x,y,t)$ is defined by convolving ommatidium output $P(x,y,t)$ with a temporal high-pass convolution kernel $H(t)$. That is,
\begin{align}
L(x,y,t) &= \int P(x,y,s)H(t-s) ds \label{LMCs-Conv} \\
H(t) &= \Gamma_{n_1,\tau_1}(t) - \Gamma_{n_2,\tau_2}(t)
\label{LMCs-HPF}
\end{align}
where $\Gamma_{n,\tau}(t)$ is a Gamma kernel, defined as
\begin{equation}
\Gamma_{n,\tau}(t) = (nt)^n \frac{\exp(-nt/\tau)}{(n-1)!\tau^{n+1}}.
\end{equation}

In the insect's visual system, before LMC relays its output to downstream neurons, it receives lateral inhibition from its adjacent neurons. In the proposed neural network, $L(x,y,t)$ is convolved with an inhibition kernel $W_1(x,y,t)$ so as to implement lateral inhibition mechanism. That is,
\begin{equation}
L_I(x,y,t) = \iiint L(u,v,s)W_1(x-u,y-v,t-s) du dv ds
\label{LMCs-Lateral-Inhibition-Mechanism}
\end{equation}
where $L_I(x,y,t)$ is the signal after lateral inhibition and $W_1(x,y,t)$ is defined by,
\begin{equation}
W_1(x,y,t) = W_{_S}^{^P}(x,y)W_{_T}^{^P}(t) +W_{_S}^{^N}(x,y)W_{_T}^{^N}(t)
\end{equation}
where $W_{_S}^{^P}(x,y)$, $W_{_S}^{^N}(x,y)$, $W_{_T}^{^P}(t)$, $W_{_T}^{^N}(t)$ are set as
\begin{align}
W_{_S}^{^{P}} &= [G_{\sigma_2}(x,y) - G_{\sigma_3}(x,y)]^+ \label{LMCs-Lateral-Inhibition-Kernel-1}\\
W_{_S}^{^{N}} &= [G_{\sigma_2}(x,y) - G_{\sigma_3}(x,y)]^- ,  \ \sigma_3 = 2 \cdot \sigma_2 \label{LMCs-Lateral-Inhibition-Kernel-2}\\
W_{_T}^{^P} &= \frac{1}{\lambda_1}\exp(-\frac{t}{\lambda_1})  \label{LMCs-Lateral-Inhibition-Kernel-3}\\
W_{_T}^{^N} &= \frac{1}{\lambda_2}\exp(-\frac{t}{\lambda_2}),  \ \lambda_2 > \lambda_1 \label{LMCs-Lateral-Inhibition-Kernel-4}.
\end{align}
where $[x]^+, [x]^-$ denote $\max (x,0)$ and $\min (x,0)$, respectively.

\subsection{Medulla Layer}
\label{Modeling-Medulla-Layer}
In the insect's visual system, medulla layer contains a great number of medulla neurons, including Tm1, Tm2, Tm3 and Mi1 \cite{behnia2014processing}. These four medulla neurons receive signals from lamina layer and respond strongly to luminance changes. More precisely, Mi1 and Tm3 neurons respond selectively to luminance increases, with the response of Mi1 delayed relative to Tm3. Conversely, Tm1 and Tm2 respond selectively to luminance decreases, with the response of Tm1 delayed relative to Tm2.

Before modeling the four medulla neurons, we first split the LMC neural outputs $L_I(x,y,t)$ into positive and negative parts denoted by $S^{^{ON}}(x,y,t)$ and $S^{^{OFF}}(x,y,t)$, respectively. That is,
\begin{align}
S^{^{ON}}(x,y,t) &= [L_I(x,y,t)]^{+} \label{ON-Channel} \\ 
S^{^{OFF}}(x,y,t) &= -[L_I(x,y,t)]^{-}  \label{OFF-Channel}
\end{align}
where $[x]^+, [x]^-$ denote $\max (x,0)$ and $\min (x,0)$, respectively. $S^{^{ON}}$ and $S^{^{OFF}}$ are also called ON and OFF signals, which are able to reflect luminance increase and decrease, respectively. 

%In the proposed feedback neural network, four medulla neurons including Tm1, Tm2, Tm3 and Mi1, not only receive feedforward signals from lamina layer, but also receive feedback signals from lobula layer (see Fig. \ref{Schematic-of-Feedback-Model}). These two signals, i.e., feedforward and feedback signals, are added together to define the outputs of four medulla neurons.

Since the Tm3 and Tm2 respond strongly to luminance increases and decreases, we use $S^{^{ON}}(x,y,t)$ and $S^{^{OFF}}(x,y,t)$
to define the outputs of Tm3 and Tm2, respectively. That is,
\begin{align}
S^{^{Tm3}}(x,y,t) = &\Big{[}\iint S^{^{ON}}(u,v,t) W_2(x-u,y-v) du dv\Big{]}^{+} \\
S^{^{Tm2}}(x,y,t) = &\Big{[}\iint S^{^{OFF}}(u,v,t) W_2(x-u,y-v) du dv\Big{]}^{+} \label{ESTMD-Tm2-Lateral-Inhibition}
\end{align}
where $S^{^{Tm3}}$ and $S^{^{Tm2}}$ denote outputs of Tm3 and Tm2 neurons, respectively; $W_2(x,y)$ is the second-order lateral inhibition kernel, defined as
\begin{equation}
W_2(x,y) = A[g(x,y)]^{+} + B[g(x,y)]^{-}
\label{ESTMD-Mdeulla-Lateral-Inhibition-Kernel-W2}
\end{equation}
where $A,B$ are constant, and $g(x,y)$ is given by
\begin{equation}
g(x,y)  = G_{\sigma_4}(x,y) - e \cdot G_{\sigma_5}(x,y) - \rho
\label{ESTMD-Mdeulla-Lateral-Inhibition-Kernel-W2-2}
\end{equation}
where $G_{\sigma}(x,y)$ is a Gaussian function and $e,\rho$ are constant.

Since the neural response of the Mi1 (or Tm1) is delayed relative to the Tm3 (or Tm2), we define the output of the Mi1 (or Tm1) using the temporally delayed output of the Tm3 (or Tm2). That is, 
\begin{align}
S^{^{Mi1}}(x,y,t) = & \int S^{^{Tm3}}(u,v,t) \cdot \Gamma_{n_{_N},\tau_{_N}}(t-s) ds \\ 
S^{^{Tm1}}(x,y,t) = & \int S^{^{Tm2}}(u,v,t) \cdot \Gamma_{n_{_F},\tau_{_F}}(t-s) ds 
\end{align}
where $S^{^{Mi1}}$ and $S^{^{Tm1}}$ represent outputs of Mi1 and Tm1, respectively; $n_{_N}, n_{_F}$ are orders of Gamma kernels while $\tau_{_N}, \tau_{_F}$ are time constants.

%\begin{equation}
%\begin{split}
%S^{^{Tm1}}(x,y,t) = & \int \Big{[}\iint S^{^{OFF}}(u,v,t) W_2(x-u,y-v) du dv\Big{]}^{+} \cdot \Gamma_{n_{_F},\tau_{_F}}(t-s) ds \\  &\cdots + {\bm{k \cdot \int F(x,y,s)\Gamma_{n_{_L},\tau_{_L}}(t-s) ds}} \\
%\end{split}
%\end{equation}

\subsection{Lobula Layer}
In the insect's visual system, STMD neurons integrate signals from medulla neurons and respond selectively to small target motion. 

In the existing ESTMD model \cite{wiederman2008model}, the output of STMD neuron $F(x,y,t)$ with visual field centered at $(x,y)$ is defined by multiplying the Tm3 neural output $S^{^{Tm3}}(x,y,t)$ with the Tm1 neural output $S^{^{Tm1}}(x,y,t)$. That is,
\begin{equation}
F(x,y,t) = S^{^{Tm3}}(x,y,t)\times S^{^{Tm1}}(x,y,t).
\label{ESTMD-Correlation}
\end{equation}

In the proposed feedback neural network, the medulla neural outputs and feedback signal are added together to define the output of the STMD neuron (see Fig. \ref{Schematic-of-Feedback-Model}). The temporally delayed model output is used as the feedback signal, which is obtained by convolving $F(x,y,t)$ with a Gamma kernel. That is, 
\begin{equation}
\begin{split}
F(x,&y,t) = \Big\{S^{^{Tm3}}(x,y,t) + \bm{k \cdot \int F(x,y,s) \cdot \Gamma_{n_{_L},\tau_{_L}}(t-s) ds} \Big\} \\
& \times \Big\{ S^{^{Tm1}}(x,y,t) + \bm{k \cdot \int F(x,y,s) \cdot \Gamma_{n_{_L},\tau_{_L}}(t-s) ds} \Big\}.
\end{split}
\label{Feedback-ESTMD-Correlation}
\end{equation}
where $n_L$  and $\tau_{_L}$ are the order and time constant of the Gamma kernel, respectively.

%\vspace{-10pt}
\section{Results and Discussions}
\label{Results-and-Discussions}

In this section, we test the ability of the proposed feedback neural network (Feedback ESTMD) for detecting small targets against cluttered backgrounds. The proposed neural network is tested on a set of image sequences produced by Vision Egg \cite{straw2008vision}. The video images are $500$ (in horizontal) by $250$ (in vertical) pixels and temporal sampling frequency is set as $1000$ Hz.

Before performing experiments, we explain how to determine the location of a small moving target using model output $F(x,y,t)$. For a given detection threshold $\gamma$, if there is a position $(x_0,y_0)$ and time $t_0$ which satisfy model output $F(x_0,y_0,t_0)>\gamma$, then we believe that a small target is detected at position $(x_0,y_0)$ and time $t_0$.
Two metrics are defined to evaluate detection performance. That is,
\begin{align}
D_R & = \frac{\text{number of true detections}}{\text{number of actual targets}} \\
F_A & = \frac{\text{number of false detections}}{\text{number of images}}
\end{align}
where $D_R$ and $F_A$ represent the detection rate and false alarm rate, respectively. The detected result is considered correct if the pixel distance between the ground truth and the result is within a threshold ($5$ pixels).

\begin{figure}[t]
	\centering
	\subfloat[]{\includegraphics[width=0.35\textwidth]{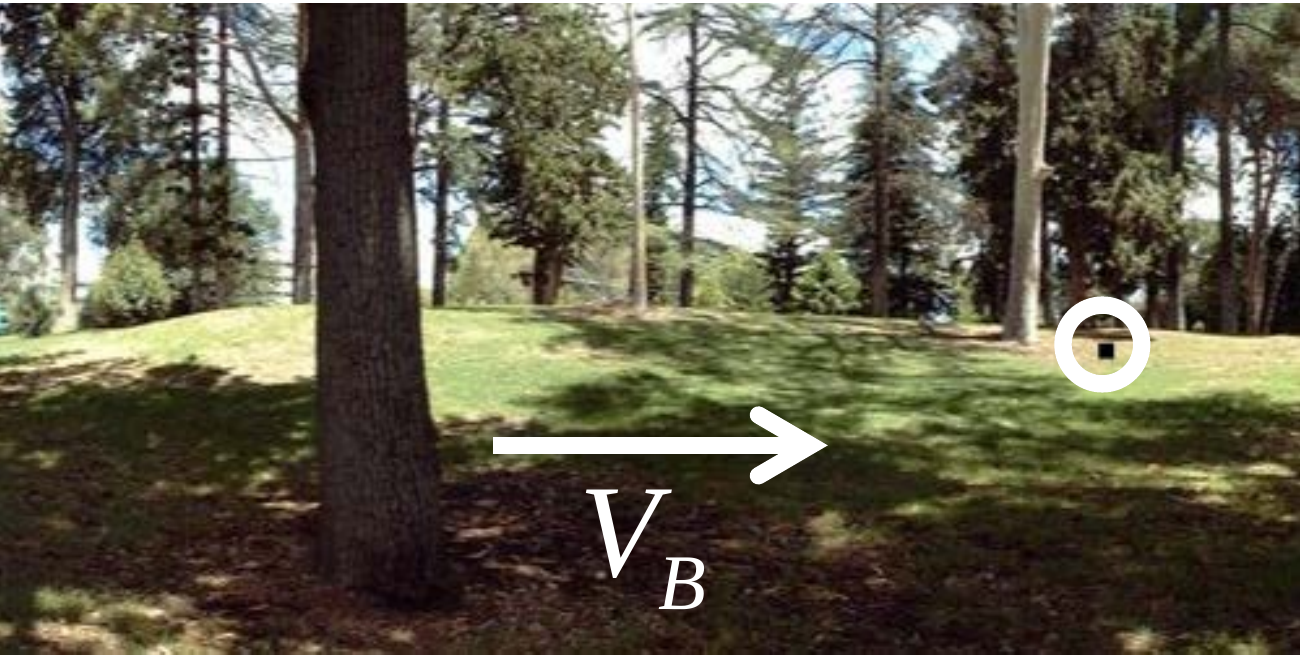}
		\label{Curvilinear-Motion-Original-Image}}
	\hfil
	\subfloat[]{\includegraphics[width=0.32\textwidth]{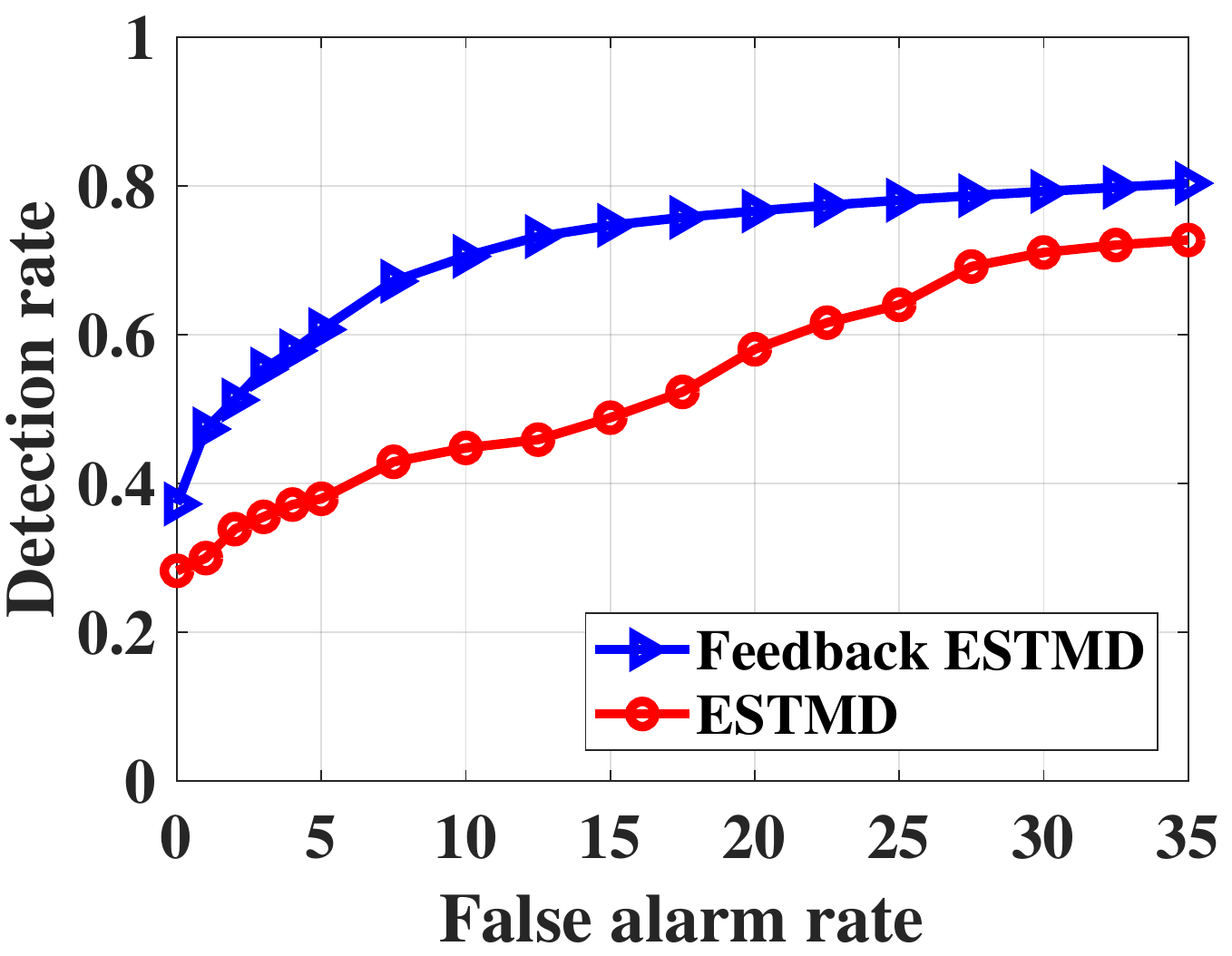}
		\label{ROC-Curve-Experiment-1}}
	\caption{(a) A representative frame of the input image sequence. The small target is highlighted by the white circle. Arrow  $V_B$ denote motion direction of the background. (b) The receiver operating characteristic (ROC) curve.}
	\label{Input-Frame-and-ROC-Curve}
\end{figure}

In the first experiment, we use an image sequence which shows a small dark target moving against the naturally cluttered background, as model input. A representative frame is shown in Fig. \ref{Input-Frame-and-ROC-Curve}(a). The background is moving from left to right and its velocity $V_B$ is set as $V_B = 250$ (pixel/second). A small target is moving against the cluttered background and its coordinate at time $t$ is set as $(500 - V_{_T} \cdot \frac{t+300}{1000}, 125+15 \cdot \sin(4\pi \frac{t+300}{1000})), t \in [0, 1000]$ ms where $V_{_T}$ denotes target velocity and is set as $V_{_T} = 500$ (pixel/second). The luminance and size of the small target are set as $50$ and $5 \times 5$ (pixel $\times$ pixel), respectively. The receiver operating characteristic (ROC) curve is presented in Fig. \ref{Input-Frame-and-ROC-Curve}(b).

Fig. \ref{Input-Frame-and-ROC-Curve}(b) is illustrating that the proposed feedback model (Feedback ESTMD) outperforms the existing model (ESTMD) at detecting small targets against naturally cluttered backgrounds. More precisely, for a given false alarm rate, feedback ESTMD has a higher detection rate than ESTMD. This also indicates that the feedback loop can improve detection performance of the existing STMD-based models.

\begin{figure}[t]
	\centering
	\subfloat[]{\includegraphics[width=0.32\textwidth]{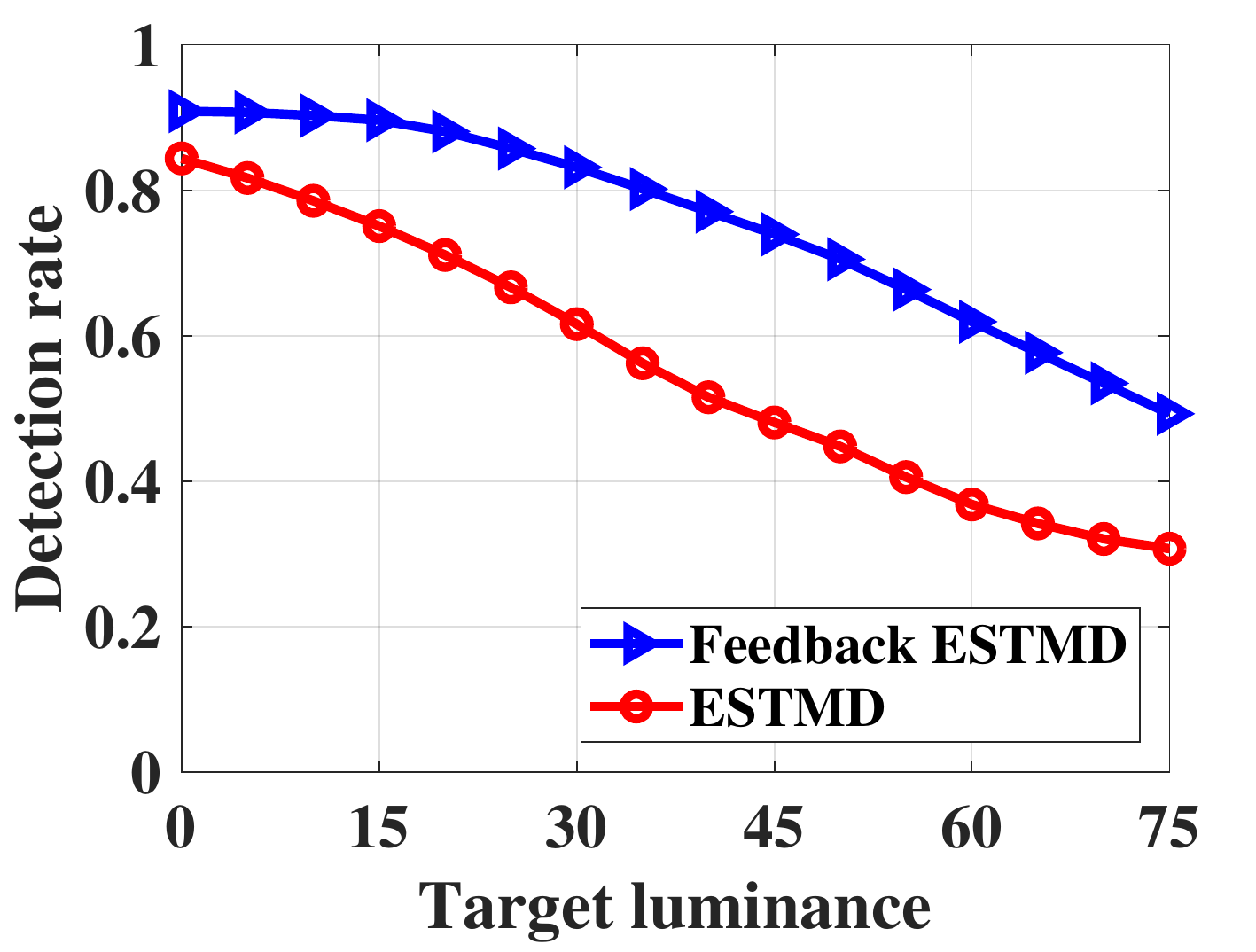}
		\label{Detection-Rate-Under-Different-Target-Luminance}}
	\hfil
	\subfloat[]{\includegraphics[width=0.32\textwidth]{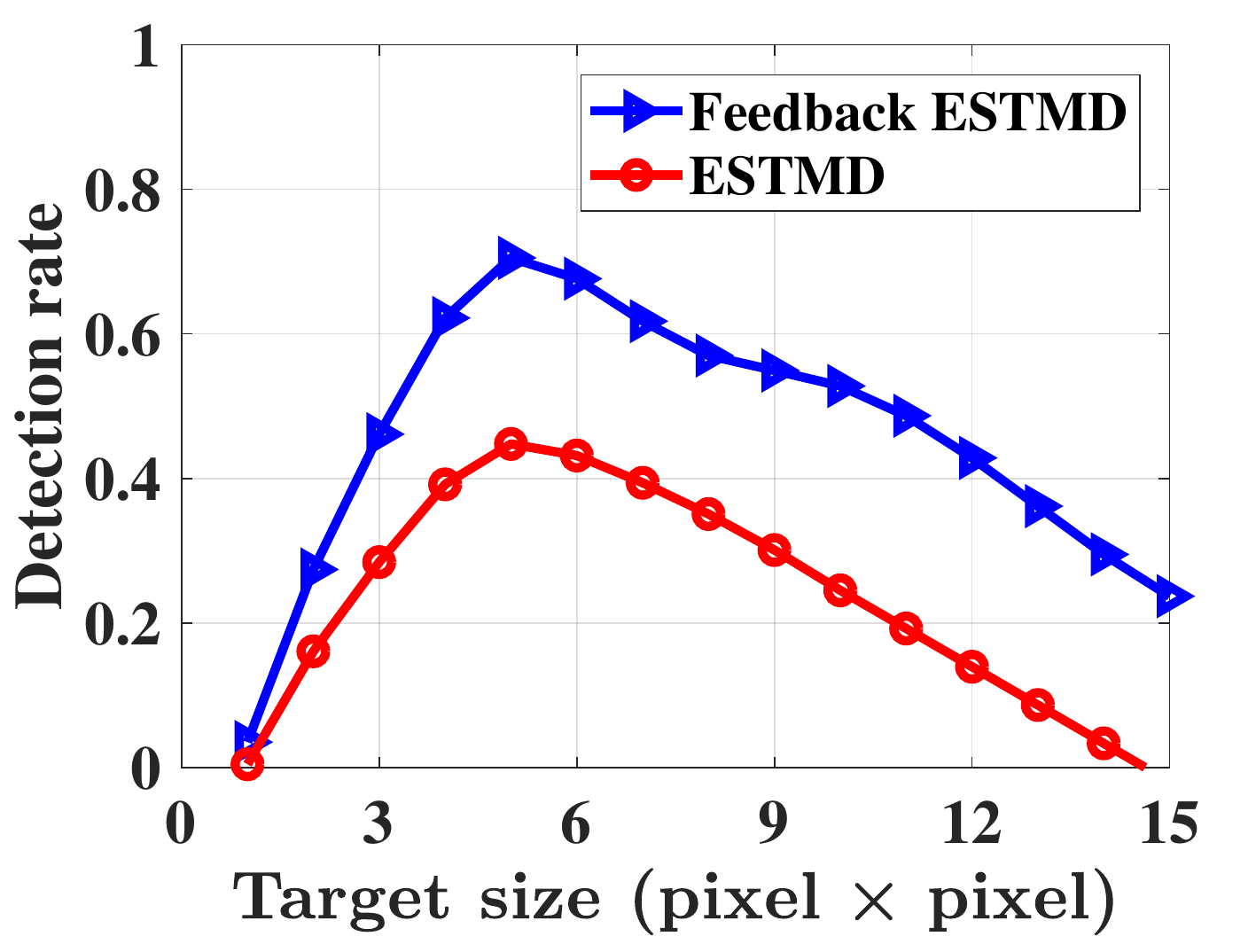}
		\label{Detection-Rate-Under-Different-Target-Size}}
	\hfil
	\subfloat[]{\includegraphics[width=0.32\textwidth]{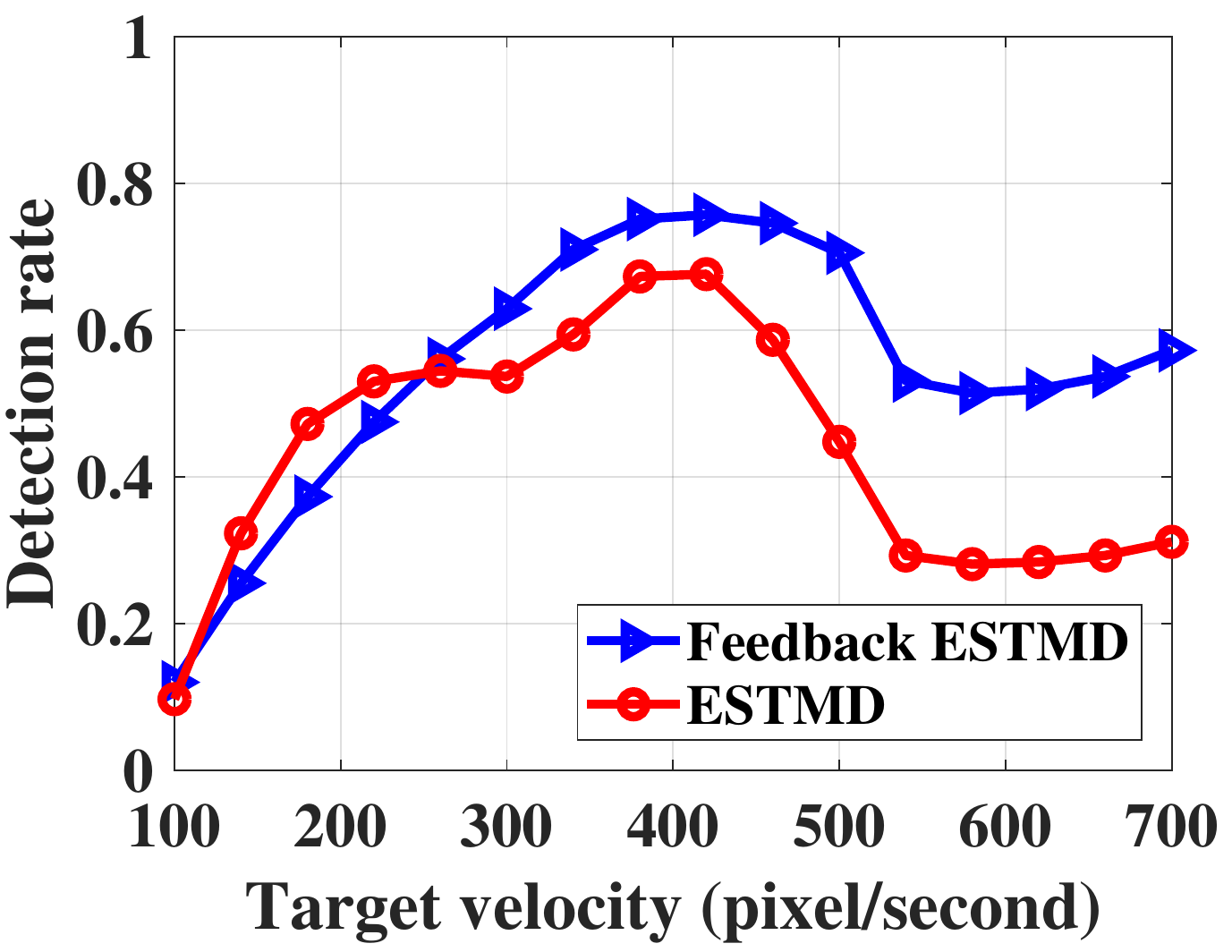}
		\label{Detection-Rate-Under-Different-Target-Velocity}}
	\hfil
	\subfloat[]{\includegraphics[width=0.32\textwidth]{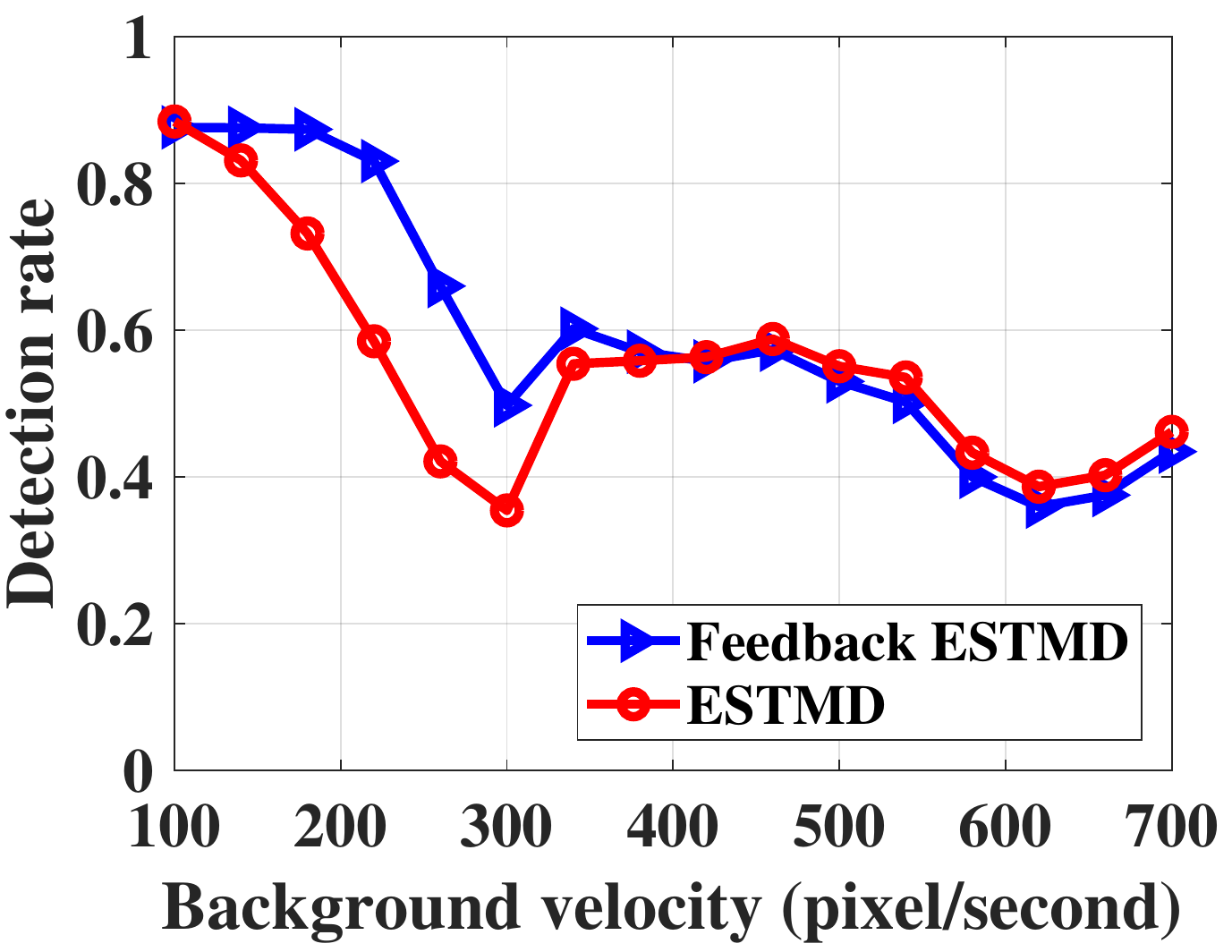}
		\label{Detection-Rate-Under-Different-Background-Velocity-Opposite-Direction}}
	\hfil
	\subfloat[]{\includegraphics[width=0.32\textwidth]{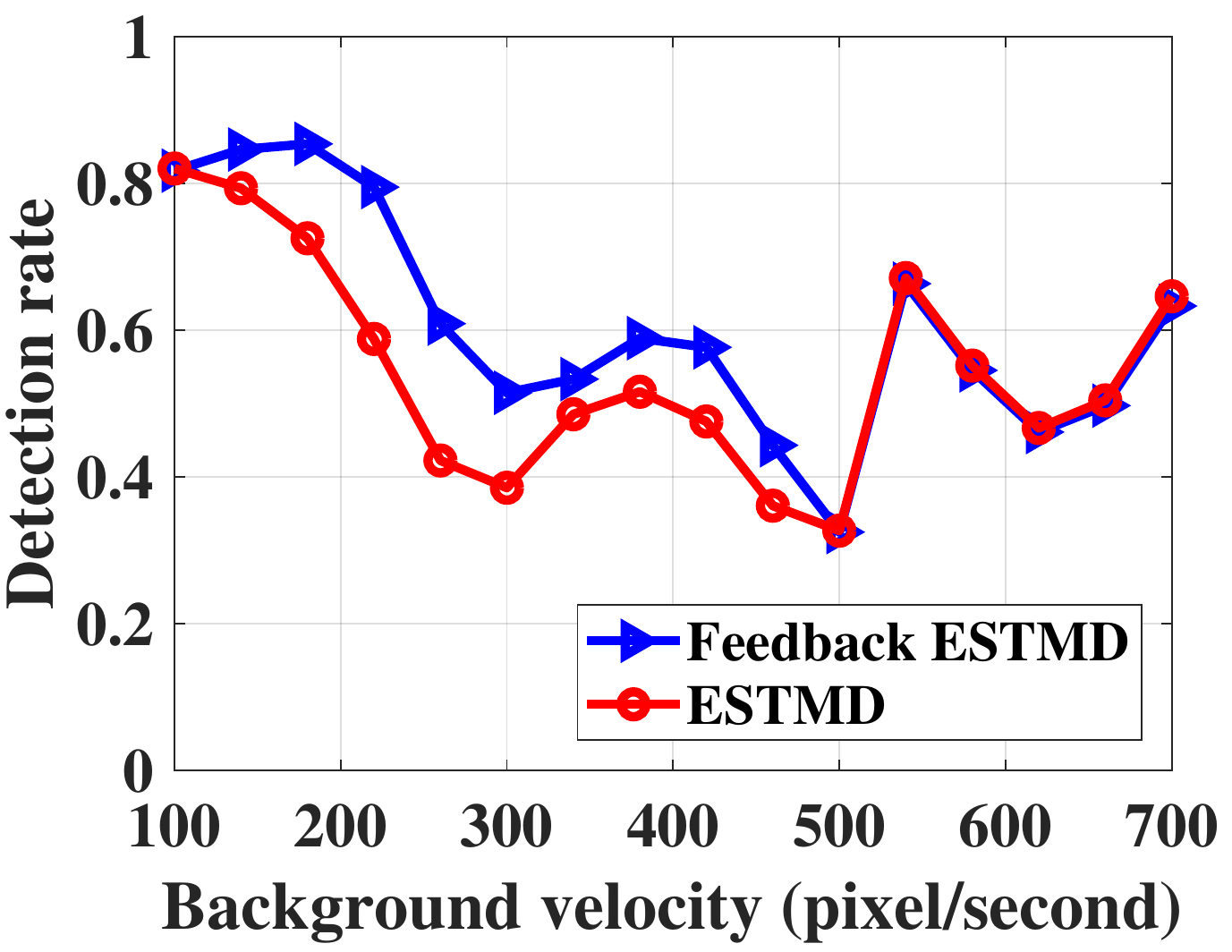}
		\label{Detection-Rate-Under-Different-Background-Velocity-Same-Direction}}
	\caption{Detection rates of the proposed feedback model (feedback ESTMD) and the existing model (ESTMD) at a fixed false alarm rate $F_A = 10$ when parameters of image sequences are changed. In each subplot, horizontal axis denotes the varying parameter while vertical axis denotes detection rate $D_R$. (a) Varying target luminance. (b) Varying target size. (c) Varying target velocity. (d) Varying background velocity when the target and the background are moving along the {\bf opposite direction}. (e) Varying background velocity when the target and the background are moving along the {\bf same direction}.}
	\label{ROC-Curve-Under-Different-Parameters}
\end{figure}

We further test these two models under different parameters of the image sequence, including target luminance, target size, target velocity, background velocity and background motion direction. In order to compare detection performances, we fix false alarm rate $F_A$ as $10$ and  illustrate detection rates of two models at this false alarm rate. The corresponding results are shown in Fig. \ref{ROC-Curve-Under-Different-Parameters}.

From Fig. \ref{ROC-Curve-Under-Different-Parameters}(a) and (b), we can see that feedback ESTMD has a better detection performance than ESTMD under different target luminance and sizes. To be more precise, the detection rate of feedback ESTMD is much higher than that of ETMD when target luminance varies (see Fig. \ref{ROC-Curve-Under-Different-Parameters}(a)). Similarly in Fig. \ref{ROC-Curve-Under-Different-Parameters}(b), the detection rate of feedback ESTMD is higher than that of ETMD under different target sizes.

From Fig. \ref{ROC-Curve-Under-Different-Parameters}(c), (d) and (e), we can find that detection performance of feedback ESTMD is dependent on velocity difference between the background and the small target. More precisely, as we can see from Fig. \ref{ROC-Curve-Under-Different-Parameters}(c), when target velocity is larger than background velocity $V_B = 250$ (pixel/second), feedback ESTMD has higher detection rates than ESTMD. However, when target velocity is smaller than background velocity,  detection rate of feedback ESTMD is slightly lower than that of ESTMD. Similar variation trend can be seen Fig. \ref{ROC-Curve-Under-Different-Parameters}(d) and (e). To be more precise, no matter whether the background and the small target are moving along the same direction or not, the detection rate of feedback ESTMD is higher than that of ESTMD when background velocity is smaller than target velocity $V_T = 500$ (pixel/second). When background velocity is larger than target velocity, detection rates of these two models show no significant difference.

\begin{figure}[t]
	%\vspace{-10pt}
	\centering
	\subfloat[]{\includegraphics[width=0.35\textwidth]{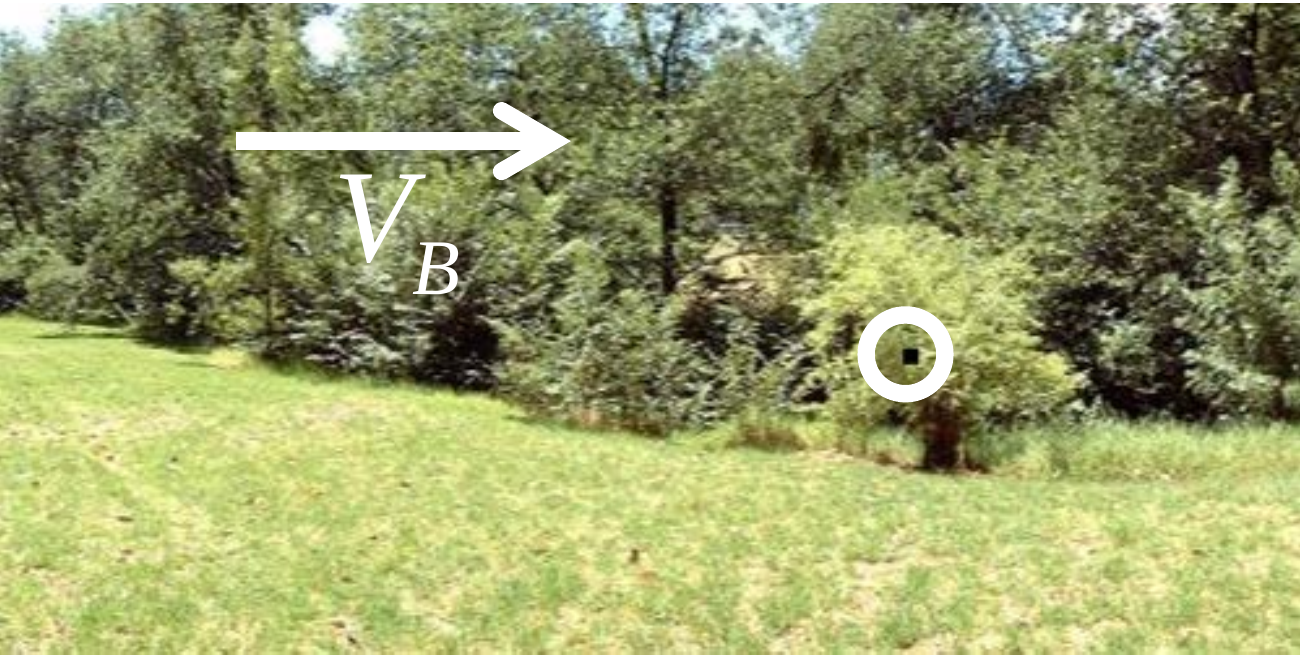}
		\label{CB-2-Frame-600}}
	\hfil
	\subfloat[]{\includegraphics[width=0.32\textwidth]{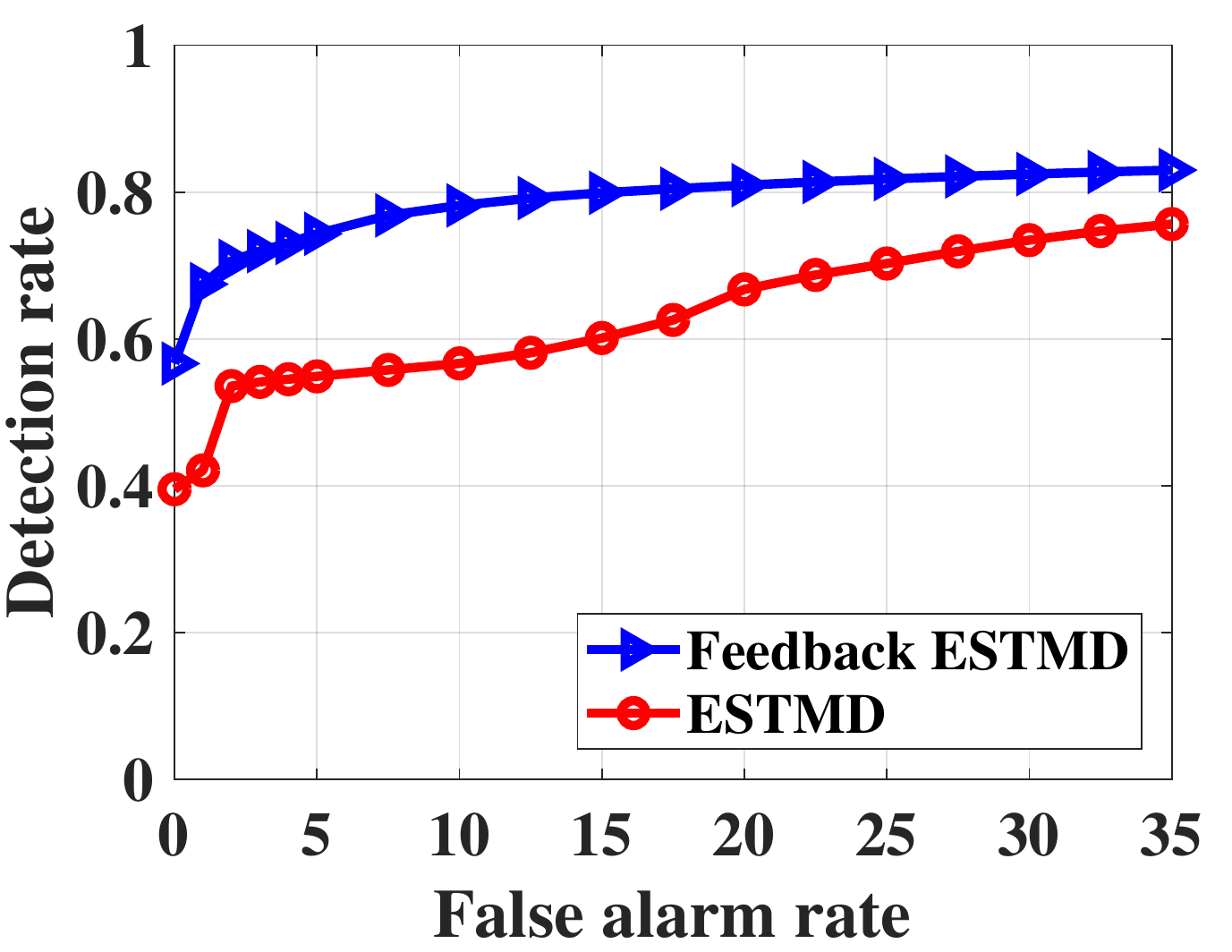}
		\label{ROC-Curve-Experiment-2}}
	\caption{(a) A representative frame of the input image sequence. The small target is highlighted by the white circle. Arrow $V_B$ denote motion direction of the background. (b) The receiver operating characteristic (ROC) curves of feedback ESTMD and ESTMD.}
	\label{Input-Frame-and-ROC-Curve-Experiment-2}
\end{figure}

\begin{figure}[t]
	%\vspace{-10pt}
	\centering
	\subfloat[]{\includegraphics[width=0.35\textwidth]{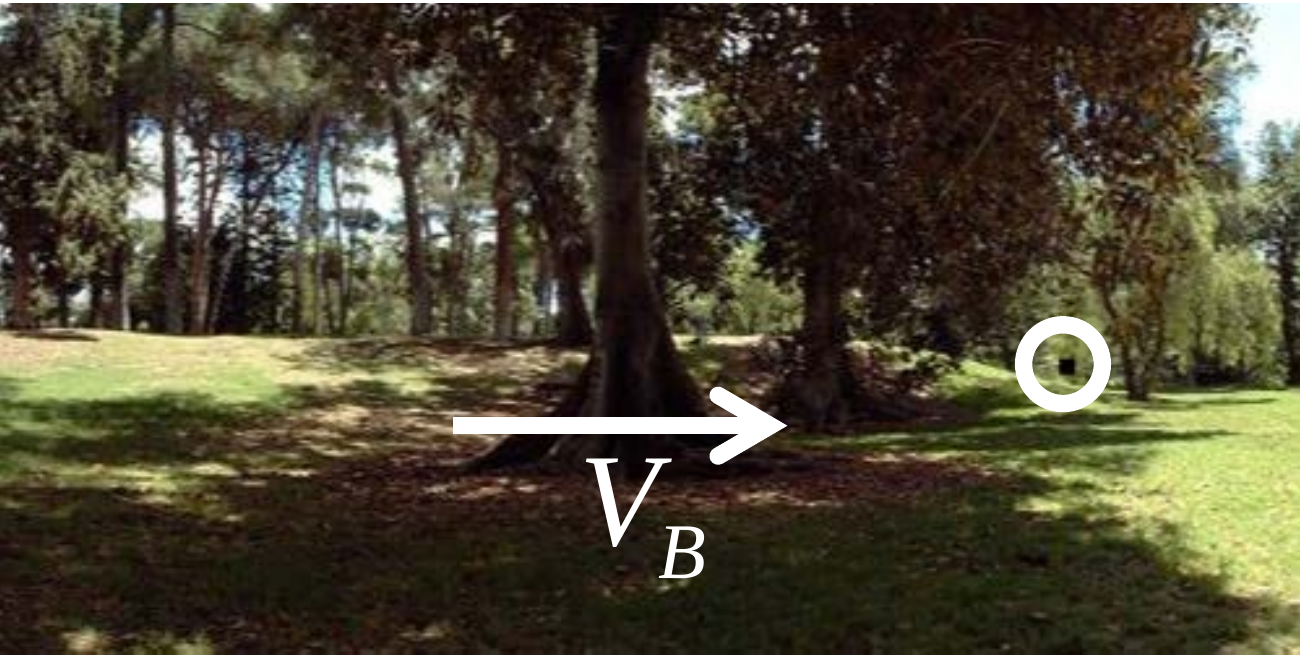}
		\label{CB-3-Frame-360}}
	\hfil
	\subfloat[]{\includegraphics[width=0.32\textwidth]{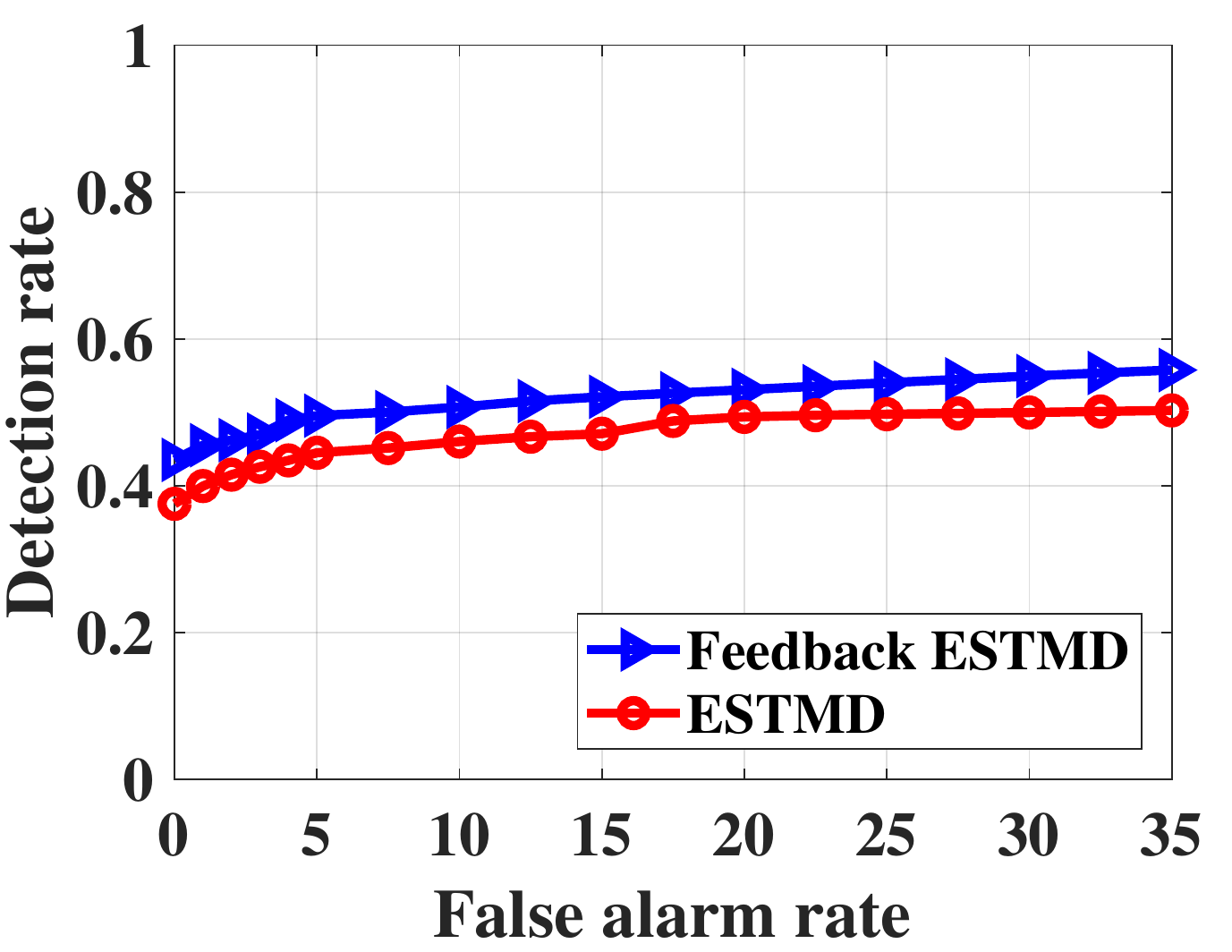}
		\label{ROC-Curve-Experiment-3}}
	\caption{(a) A representative frame of the input image sequence. The small target is highlighted by the white circle. Arrow $V_B$ denote motion direction of the background. (b) The receiver operating characteristic (ROC) curves of feedback ESTMD and ESTMD.}
	\label{Input-Frame-and-ROC-Curve-Experiment-3}
\end{figure}

In the second and third experiment, we test the proposed feedback model in different cluttered backgrounds. Two image sequences with different backgrounds are used as model input in these two experiments. Two representative frames are presented in Fig. \ref{Input-Frame-and-ROC-Curve-Experiment-2}(a) and Fig. \ref{Input-Frame-and-ROC-Curve-Experiment-3}(a), respectively. In these two image sequences, backgrounds are all moving from left to right and their velocities are set as $250$ (pixel/second). A small target whose luminance, size are set as $50$ and $5 \times 5$ (pixel $\times$ pixel), is moving against cluttered backgrounds. The coordinate of the small target at time $t$ equals to $(500 - V_{_T} \frac{t+300}{1000}, 125+15 \cdot \sin(4\pi \frac{t+300}{1000})), t \in [0, 1000]$ ms where $V_{_T}$ is set as $500$ (pixel/second).

From Fig. \ref{Input-Frame-and-ROC-Curve-Experiment-2}(b) and Fig. \ref{Input-Frame-and-ROC-Curve-Experiment-3}(b), we can see that feedback ESTMD  has a better performance than ESTMD. For a given false alarm rate, the detection rate of feedback ESTMD is higher than that of ESTMD. This indicate that feedback ESTMD performs better than ESTMD in different cluttered backgrounds.

\vspace{-10pt}
\section{Conclusion}
\label{Conclusion}
In this paper, we proposed a feedback neural network for small target detection against naturally cluttered backgrounds. In order to  form a feedback loop, network output is temporally delayed and then relayed to middle neural layer as feedback signal. Feedback signal is added on outputs of middle neural layer for weakening responses to background noise. Systematic experiments showed that the proposed feedback neural network has a much better performance than the existing ESTMD model, if there is velocity difference between the background and the small target. In the future, we will further combine feedback loops with visual attention mechanisms for improving detection performances of models.

\vspace{-10pt}
\subsubsection*{Acknowledgments.}
This research was supported by EU FP7 Project HAZCEPT (318907), HORIZON 2020 project STEP2DYNA (691154), ENRICHME (643691) and the National Natural Science Foundation of China under the grant no. 11771347.

%Author, F.: Article title. Journal \textbf{2}(5), 99--110 (2016)
%Author, F., Author, S.: Title of a proceedings paper. In: Editor,
%F., Editor, S. (eds.) CONFERENCE 2016, LNCS, vol. 9999, pp. 1--13.
%Springer, Heidelberg (2016). \doi{10.10007/1234567890}
%Author, F., Author, S., Author, T.: Book title. 2nd edn. Publisher,
%Author, A.-B.: Contribution title. In: 9th International Proceedings
%on Proceedings, pp. 1--2. Publisher, Location (2010)
%LNCS Homepage, \url{http://www.springer.com/lncs}. Last accessed 4
%\end{thebibliography}

\bibliographystyle{splncs04}
\bibliography{reference.bib}

\end{document}